\documentclass[sigconf]{acmart} 

\usepackage{xcolor}
\usepackage{amsmath}
\usepackage{subcaption}
\DeclareMathOperator*{\argmax}{arg\,max}

\AtBeginDocument{%
  \providecommand\BibTeX{{%
    \normalfont B\kern-0.5em{\scshape i\kern-0.25em b}\kern-0.8em\TeX}}}

\copyrightyear{2022}
\acmYear{2022}
\setcopyright{rightsretained}
\acmConference[WWW '22]{Proceedings of the ACM Web Conference 2022}{April 25--29, 2022}{Virtual Event, Lyon, France}
\acmBooktitle{Proceedings of the ACM Web Conference 2022 (WWW '22), April 25--29, 2022, Virtual Event, Lyon, France}\acmDOI{10.1145/3485447.3512134}
\acmISBN{978-1-4503-9096-5/22/04}

\begin{document}

\newcommand{\Sref}[1]{\S\ref{#1}}
\newcommand{\fref}[1]{figure~\ref{#1}}
\newcommand{\Fref}[1]{Figure~\ref{#1}}
\newcommand{\tref}[1]{table~\ref{#1}}
\newcommand{\Tref}[1]{Table~\ref{#1}}
\newcommand{\Aref}[1]{Appendix~\ref{#1}}

\title{Controlled Analyses of Social Biases in Wikipedia Bios}


\author{Anjalie Field}
\email{anjalief@cs.cmu.edu}
\orcid{0000-0002-6955-746X}
\affiliation{
\institution{Carnegie Mellon University}
\country{USA}
}

\author{Chan Young Park}
\email{chanyoun@cs.cmu.edu}
\orcid{0000-0002-2720-0112}
\affiliation{%
  \institution{Carnegie Mellon University}
    \country{USA}
}

\author{Kevin Z. Lin}
\email{kevinL1@wharton.upenn.edu}
\orcid{0000-0002-1236-9847}
\affiliation{%
  \institution{University of Pennsylvania}
    \country{USA}
}

\author{Yulia Tsvetkov}
\email{yuliats@cs.washington.edu}
\orcid{0000-0002-4634-7128}
\affiliation{%
  \institution{University of Washington}
  \country{USA}
}


\begin{abstract}
Social biases on Wikipedia, a widely-read global platform, could greatly influence public opinion.
While prior research has examined man/woman gender bias in biography articles, possible influences of other demographic attributes limit conclusions.
In this work, we present a methodology for analyzing Wikipedia pages about people that isolates dimensions of interest (e.g., gender), from other attributes (e.g., occupation). Given a target corpus for analysis (e.g.~biographies about women), we present a method for constructing a comparison corpus that matches the target corpus in as many attributes as possible, except the target one.
We develop evaluation metrics to measure how well the comparison corpus aligns with the target corpus and
then examine how articles about gender and racial minorities (cis. women, non-binary people, transgender women, and transgender men; African American, Asian American, and Hispanic/Latinx American people) differ from other articles.
In addition to identifying suspect social biases, our results show that failing to control for covariates can result in different conclusions and veil biases.
Our contributions include methodology that facilitates further analyses of bias in Wikipedia articles, findings that can aid Wikipedia editors in reducing biases, and a framework and evaluation metrics to guide future work in this area.
\end{abstract}



\maketitle

\section{Introduction}
\label{sec:introduction}

Since its inception, Wikipedia has attracted the interest of researchers in various disciplines because of its unique community and departure from traditional encyclopedias \cite{jullien2012we,kolbitsch2006transformation}.
The collaborative knowledge platform allows for fast and inexpensive dissemination of information, but it risks introducing social and cultural biases \cite{kolbitsch2006transformation}.
These biases can influence readers and be absorbed and amplified by computational models, as Wikipedia has become a popular data source  \cite{bolukbasi2016man,zhao2017men,peters2018deep,mora2019systematic,redi2020taxonomy}.
Considering the large volume of Wikipedia data, automated methods are vital for identifying social and cultural biases on the platform. 
In this work, we develop methodology to identify \textit{content bias}: systemic differences in the text of articles about people with different demographic traits.

Prior computational work on Wikipedia bias has primarily focused on binary gender and examined differences in articles about men and women, e.g., pages for women discuss personal relationships more often than pages for men \cite{graells2015first,wagner2016women,adams2019counts}. 
However, attributes about people other than their gender limit conclusions that can be drawn from these analyses.
For example, there are more male than female athletes on Wikipedia, so it is difficult to disentangle if differences occur because women and men are presented differently, or because non-athletes and athletes are presented differently \cite{graells2015first,wagner2016women,hollink2018using}. Existing work has incorporated covariates as explanatory variables in a regression model, which restricts analysis to regression models and requires enumerating all attributes \cite{wagner2016women}.

In contrast, we develop a \textit{matching algorithm} that enables analysis by isolating target dimensions (\Sref{sec:method}). Given a corpus of Wikipedia biography pages that contain \textit{target} attributes (e.g.~pages for cisgender women), our algorithm builds a matched \textit{comparison corpus} of biography pages that do not (e.g.~pages for cis.~men).
We construct this corpus to closely match the target corpus on all known attributes except the targeted one (e.g.~gender) by using pivot-slope TF-IDF weighted attribute vectors.
Thus, examining differences between the two corpora can reveal \textit{content bias} \cite{young2016s} related to the target attribute.
We develop frameworks to evaluate our methodology that measure how closely constructed comparison corpora match simulated target corpora (\Sref{sec:results}).

We ultimately use this method to analyze biography pages that Wikipedia editors or readers may perceive as describing gender (cis.~women, non-binary people, transgender women, and transgender men) and racial (African American, 
Asian American, Hispanic/Latinx American) minorities (\Sref{sec:analysis}). 
We additionally intersect these dimensions and examine portrayals of African American women~\cite{crenshaw1989demarginalizing}.
Our analysis focuses on statistics that have been used in prior work to assess article quality, including article lengths, section lengths, and edit counts on English Wikipedia. We also consider language availability and length statistics in other language editions \cite{wagner2015s}. This analysis reveals systemic differences: for example, articles about cis.~women tend to be shorter and available in fewer languages than articles about cis.~men; articles about African American women tend to be available in more languages than comparable other American men, but fewer languages than comparable other American women. 
Disparities can be signs of editor bias, suggesting that Wikipedia editors write articles about people they perceive as gender or racial minorities differently from other articles, or they can be signs of imbalances in society that are reflected on Wikipedia. While we cannot always distinguish these sources, where possible, we target editor bias, which editors can investigate and mitigate.

To the best of our knowledge, this is the first work to examine gender disparities in Wikipedia biography pages beyond cis.~women, the first large-scale analysis of racial disparities \cite{adams2019counts}, and the first consideration of intersectionality in Wikipedia biography pages.
Overall, our work offers methodology and initial findings for uncovering content biases on Wikipedia, as well as a framework and evaluation metrics for future work in this area.

\section{Related Work}
\label{sec:related_work}

Systemic differences in Wikipedia coverage of different types of people can under-serve or unintentionally influence readers and can perpetuate bias and stereotypes in machine learning models and social science studies that rely on Wikipedia data \citep{zhao2017men,Olteanu2019,blodgett-etal-2020-language,field-etal-2021-survey,redi2020taxonomy}.
Most prior work on Wikipedia bias focuses on gender and  \textit{coverage bias}, \textit{structural bias}, or \textit{content bias}.
\textit{Coverage bias} involves examining how likely notable men and women are to have Wikipedia articles
and on article length \cite{reagle2011gender,wagner2015s,young2016s}.
On average, articles about women are longer than articles about men \cite{graells2015first,reagle2011gender,wagner2015s,young2016s}.
\textit{Structural bias} denotes differences in article meta-properties such as links between articles, diversity of citations, and edit counts \cite{young2016s,wagner2015s}. Examinations have found structural bias against women (e.g.~all biography articles tend to link to articles about men more than women) \cite{young2016s,wagner2015s,wagner2016women,eom2015interactions}. Finally, \textit{content bias} considers article text itself. Latent variable models  \cite{bamman2014unsupervised}, lexicon counts, and pointwise mutual information (PMI) scores \cite{graells2015first,wagner2016women} have suggested that pages for women discuss personal relationships more frequently than pages for men.
In the past, research on biases in Wikipedia has drawn the attention of the editor community and led to changes on the platform \cite{reagle2011gender,langrock2020gender}, which could explain why similar studies sometimes have different findings and also motivates our work.

However, many studies draw limited conclusions because of data confounds. For example, word statistics suggest that the most over-represented words on pages for men as opposed to women are sports terms: ``footballer'', ``baseball'', ``league'' \cite{graells2015first,wagner2016women}. This result is not necessarily indicative of \textit{bias}; Wikipedia editors do not omit the football achievements of women. Instead, in society and on Wikipedia, there are more male football players than female ones. Thus, these word statistics likely capture imbalances in occupation, rather than gender. While gender imbalance in sports and the acceptance among editors that football players are notable enough to merit Wikipedia articles are worth studying in their own right, in our setting, they limit our ability to examine how editors might write articles about men and women differently. Some traits can be accounted for, e.g.~using explanatory variables in a regression model \cite{wagner2016women}, but it is difficult to explicitly enumerate all possible traits and limits analysis to particular models.

These traits also impact analyses of different language editions, which often have focused on ``local heros'' and how language editions tend to favor people whose nationality is affiliated with the language, in terms of article coverage, length, and visibility \cite{eom2015interactions,callahan2011cultural,hecht2010tower}.
In investigating cross-lingual gender bias, \citet{hollink2018using} suggest that their findings are influenced by nationality and birth year more than by gender, demonstrating how the ``local heros'' effect can complicate analysis.
Language editions can also have systemic differences due to differing readership and cultural norms.
These differences are often justified, since language editions serve different readers \cite{callahan2011cultural,hecht2010tower}, but 
can confound research questions.
Do English biographies about women contain more words related to family than Spanish ones because there is greater gender bias in English \cite{wagner2015s}?
Or because English editors generally discuss family more than Spanish ones, regardless of gender?
Comparing biography pages of men and women in each language partially alleviates this ambiguity but other factors may also be influential \cite{konieczny2018gender}.


While our work focuses on analyzing biases in Wikipedia articles regardless of their origin, we briefly discuss possible sources of bias as motivation.
First, surveys have suggested the Wikipedia editor community lacks diversity and is predominately white and male \cite{kolbitsch2006transformation,hill2013wikipedia,Shaw2018},\footnote{\url{https://meta.wikimedia.org/wiki/Research:Wikipedia\_Editors\_Survey\_2011\_April}}
though given efforts to improve diversity, editor demographics may have changed in the last decade \cite{konieczny2018gender,lam2011wp,eckert2013wikipedia,collier2012conflict}. 
Second, bias can propagate from \textit{secondary sources} that  editors draw from in accordance with Wikipedia's ``no original research'' policy \cite{luo2018ladies}. Finally, bias on Wikipedia may reflect broader societal biases. For example, women may be portrayed as less powerful than men on Wikipedia because editors write imbalanced articles, because other coverage of women such as newspaper articles downplay their power, or because societal constraints prevent women from obtaining the same high-powered positions as men \cite{adams2019counts}.

Finally, nearly all of the cited work focuses on men/women gender bias. Almost no computational research has examined bias in Wikipedia biographies at-scale along other dimensions, even though observed racial bias has prompted edit-a-thons to correct omissions.\footnote{\url{https://en.wikipedia.org/wiki/Racial_bias_on_Wikipedia}} Two notable exceptions: \citet{adams2019counts} examine gender and racial biases in pages about sociologists and \citet{park2021multilingual} examine disparate connotations in pages about LGBT people.

\section{Methodology}
\label{sec:method}

\subsection{Matching Methodology}
\label{sec:matching_method}

We present a method for identifying a \textit{comparison} biography page for every page that aligns with a target attribute, where the comparison page closely matches the target page on all known attributes except the target one.\footnote{All code and data is available at 
\url{https://github.com/anjalief/wikipedia_bias_public}
}

The concept originates in adjusting observational data to replicate the conditions of a randomized trial; from the observational data, researchers construct treatment and control groups so that the distribution of all covariates except the target one is as identical as possible between the two groups \cite{rosenbaum1983central}.\footnote{We use target/comparison instead of treatment/control to clarify that our work does not involve any actual ``treatment''.}
Then by comparing the constructed treatment and control groups, researchers can isolate the effects of the target attribute from other confounding variables. Matching is also gaining attention in language analysis \cite{Choudhury2016, chandrasekharan2017you, egami2018make, roberts2018adjusting, keith2020text}.
Here, our target attribute is gender or race as likely to be perceived by editors and readers. We aim to create corpora that balance other characteristics, such as age, occupation, and nationality, that could affect how articles are written.

Given a set of target articles $\mathcal{T}$ (e.g.~all biographies about women), our goal is to construct a set of comparison articles $\mathcal{C}$  from a set of candidates $\mathcal{A}$ (e.g.~all biographies about men), such that $\mathcal{C}$ has a similar covariate distribution as $\mathcal{T}$ for all covariates except the target attribute.
We construct $\mathcal{C}$  using greedy matching.
For each $t {\in} \mathcal{T}$, we identify $c_{best} {\in} \mathcal{A}$ that best matches $t$ and add $c_{best}$ to $\mathcal{C}$. If $t$ is about an American female actress, $c_{best}$ may be about an American male actor. To identify $c_{best}$, we leverage the category metadata associated with each article. For example, the page for Steve Jobs includes the categories ``Pixar people'', ``Directors of Apple Inc.'', ``American people of German descent'', etc. While articles are not always categorized correctly or with equal detail, using this metadata allows us to focus on covariates that are likely to reflect the way the article is written. People may have relevant traits that are not listed on their Wikipedia page, but if no editor assigned a category related to the traits, we have no reason to believe editors were aware of them nor that they influenced edits. We describe 6 methods for identifying $c_{best} {\in} A$. $CAT(c)$ denotes the set of categories associated with $c$.

\textbf{\textsc{Number}} We choose $c_{best}$ as the article with the most number of categories in common with $t$, which is intuitively the best match.

\textbf{\textsc{Percent}} \textsc{Number} favors articles with more categories. For example, a candidate $c_i$ that has 30 categories is more likely to have more categories in common with $t$ than a candidate $c_j$ that only has 10 categories. However, this does not necessarily mean that $c_i$ has more traits in common with the person $t$---it suggests that the article is better written. We can reduce this favoritism by normalizing the number of common categories by the total number of categories in the candidate $c_i$, i.e.
$c_{best} = \argmax_{c_{i}} | CAT(c_i) \cap CAT(t) | \frac{1}{|CAT(c_i)|} $

\textbf{\textsc{TF-IDF}} Both prior methods assume that all categories are equally meaningful, but this is an oversimplification. A candidate $c_i$ that has ``American short story writers'' in common with $t$ is more likely to be a good match than one with ``Living People'' in common. We use TF-IDF weighting to up-weight rarer categories \cite{salton1988term}. We represent each $c_i {\in} \mathcal{A}$ as a sparse category vector, where each element is a product between the frequency of the category in $c_i$, ($\frac{1}{|CAT(c_i)|}$ if the category is in $CAT(c_i)$, 0 otherwise) and the inverse frequency of the category, which down-weights broad common categories. We select $c_{best}$ as the $c_i$ with the highest cosine similarity between its vector and a similarly constructed vector for $t$.

\textbf{\textsc{Propensity}} For each article we construct a propensity score, an estimate of the probability that the article contains the target attribute \citep{rosenbaum1983central,Rubin85}, using a logistic regression classifier trained on one-hot-encoded category features.
We then choose $c_{best}$ as the article the closest propensity score  to $t$'s.
Propensity matching is not ideal in our setting,
first, because it was designed for lower-dimensional covariates and has been shown to fail with high-dimensional data, and second, because it does not necessarily produce matched pairs that are meaningful, which precludes manually examining matches \cite{roberts2018adjusting}. Nevertheless, we include it as a baseline, because it is a popular method for controlling for confounding variables.

\textbf{\textsc{TF-IDF Propensity}}
We construct an additional propensity score model, where we use TF-IDF weighted category vectors as features instead of one-hot encoded vectors.

\textbf{\textsc{Pivot-Slope TF-IDF}} \textsc{TF-IDF} and \textsc{Percent} both include the term $\frac{1}{|CAT(c_i)|}$ to normalize for articles having different numbers of categories. However, information retrieval research suggests that it over-corrects and causes the algorithm to favor articles with fewer categories \cite{singhal2017pivoted}. Instead, we adopt pivot-slope normalization \cite{singhal2017pivoted} and normalize TF-IDF terms with an adjusted value:
$(1.0 - slope) * pivot + slope * |CAT(c_i)|$.
This approach requires setting the slope and the pivot, which control the strength of the adjustment. Following \citet{singhal2017pivoted}, we set the pivot to the average number of categories across all articles, and tune the slope over a development set. Tuning the slope is important, as changing the parameter does change the selected matches. \textsc{Pivot-Slope TF-IDF} is our novel proposed approach.

In practice, it is likely not possible to identify close matches for every target article, i.e. there may be characteristics of people in the target corpus that are not shared by anyone in the comparison corpus. To account for this, we discard ``weak matches'': for direct matching methods, pairs with $<2$ categories in common, for propensity matching methods, pairs whose difference in propensity scores is $>1$ standard deviation away from the mean difference.
We provide additional details on experimental setups in \Aref{sec:app_experimental_setup}.

\subsection{Model Assumptions and Limitations}
\label{sec:limitations}

In this section, we clarify how some of the assumptions required by our methodology limit the way it should be used.
First, our matching method depends on categories, which are imperfect controls. While we take some steps to account for this, e.g., excluding articles with few categories, systemic differences in category tagging could reduce the reliability of matching (we did not observe evidence of this). Using categories as covariates also precludes us from identifying systemic differences in how article categories are assigned. Instead our work focuses on differences in articles, given the current category assignments.
Second, our methodology is only meaningful where it is possible to establish high-quality matches. If there are people with characteristics in our target corpus that do not present in our comparison corpus, we cannot draw controlled comparisons. In practice, we operationalize this limitation by discarding pairs of articles that are poorly matched (described in \Sref{sec:matching_method}) and only computing analyses over sets of articles where there is covariate overlap.
Third, we note that while we borrow some methodology and terminology from causal inference, our setup is not conducive to a strictly causal framework, and we do not suggest that all results imply causal relations. As discussed in \Sref{sec:related_work}, it is difficult to determine if article imbalances are the result of Wikipedia editing, societal biases, or other factors, meaning there are confounding variables we cannot control for.
To summarize, we aim to identify systemic differences in articles about different groups of people, and we view the main use case of our model as identifying sets of articles that may benefit from manual investigation and editing.

\subsection{Evaluation Framework}
\label{sec:experiments}

We devise schemes to evaluate both how well each matching metric creates comparison groups with similar attribute distributions as target groups and if the metric introduces imbalances, e.g.~by favoring articles with fewer categories. Given matched target and comparison sets, we assess matches using several metrics:

\textbf{Standardized Mean Difference (SMD)} SMD (the difference in means between the treatment and control groups divided by the pooled standard deviation) is the standard method used to evaluate covariate balance after matching \cite{Zhang2019}. We treat each category as a binary covariate that can be present or absent for each article. We then compute SMD for each category and report the average across all categories (AVG SMD) as well as the percent of categories where SMD$>$0.1. There is no widely accepted standardized bias criterion, but prior work suggests 0.25 or 0.1 as reasonable cut-offs \citep{harder2010propensity}.
High values indicate that the distribution of categories is very different between the target and the comparison groups.

\textbf{Number of Categories} As discussed in \Sref{sec:method}, the proposed methods may favor articles with more or fewer categories. Thus, we compute the SMD between the number of categories in the target group and comparison groups. A high value indicates favoritism.

\textbf{Text Length} The prior two metrics focus on the categories, but categories are a proxy for confounds in the text. We ultimately seek to assess how well matching controls for differences in the actual articles. We first compare article lengths (word counts) using SMD.

\textbf{Polar Log-odds (PLO)}  We use log-odds with a Dirichlet prior \cite{monroe2008fightin} to compare vocabulary differences between articles, where high log-odds polarities indicate dissimilar vocabulary.\footnote{Details on PLO and KL Divergence are provided in \Aref{sec:app_experimental_setup}}

\textbf{KL Divergence} Beyond word-level differences, we compute the KL-divergence between 100-dimensional topic vectors derived from an LDA model \citep{blei2003latent} in both the target-comparison (KL) and the comparison-target directions (KL 2).

We compute these metrics over three types of target sets:

\textit{Article-Sampling} We construct simulated target sets by randomly sampling 1000 articles. Because we do not fix a target attribute, we expect a high quality matching algorithm to identify a comparison set that matches very closely to the target set, without creating imbalances, e.g. by favoring longer articles with more categories.

\textit{Category-Sampling} We randomly sample one category that has at least 500 members, and then sample 500 articles from the category. We do not expect there to be any bias towards a single category, since most categories are very specific, e.g.~``Players of American football from Pennsylvania''. While articles for football players might have different characteristics than other articles, we would not expect articles for players from Pennsylvania to be substantially different than articles for players from New York or New Jersey. Thus, as in the article-sampling setup, we can evaluate both attribute distributions and artificial imbalances. However, this setup more closely replicates the intended analysis, as we ensure that all people in the target group have a common trait.

\textit{Attribute-specific} We evaluate how well each method balances covariates in our analysis setting, e.g., when comparing articles about women and men. In this setting, we only consider how well the method balances covariates (SMD), using heuristics to exclude categories that we expect to differ between groups (e.g., when comparing cis. men and women, we exclude categories containing the word ``women'').
We cannot examine other criterion, such as text length, because we cannot distinguish if differences between the target and comparison sets are signs of social bias or poor matching, especially considering prior work has suggested text length differs for people of different genders  \cite{graells2015first,reagle2011gender,wagner2015s,young2016s}. Instead, we use the synthetically constructed \textit{article-sampling} and \textit{category-sampling} to examine signs of favoritism in the algorithm and how well the matching controls for confounds in the article text.

\begin{figure*}
    \centering
    \includegraphics[width=\textwidth]{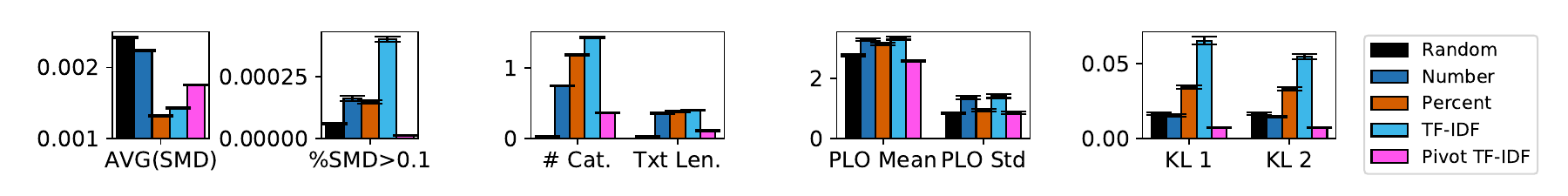}
    \includegraphics[width=\textwidth]{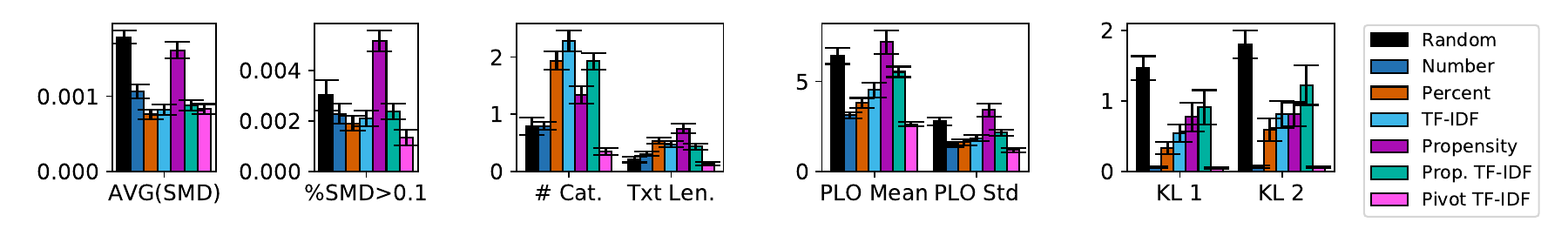}
       \Description[Bar graphs of matching evaluation metrics]{Bar graphs of evaluation metrics in two settings for seven matching methods, showing propensity TF-IDF generally outperforms other methods}
    \caption{Evaluation of matching methods using article-sampling (top) and category-sampling (bottom), with 99\% confidence intervals computed over 100 simulations. Lower scores indicate better matches; pivot-slope TF-IDF performs best overall. We omit propensity matching in article-sampling, as it is not meaningful.}
    \label{fig:random_eval}
\end{figure*}

\subsection{Data}
\label{sec:data}

We gathered all articles with the category ``Living people'' on English Wikipedia in March 2020. We discarded articles with ${<2}$ categories, ${<100}$ tokens, or marked as stubs (containing a category like ``Actor stubs''). We use English categories for matching, which we expect to be the most reliable, because English has the most active editor community. We ignore categories focused on article properties instead of people traits using a heuristics, e.g., categories containing ``Pages with''.
Our final corpus consists of 444,045 articles, containing 9.3 categories and 628.2 tokens on average. The total number of categories considered valid for matching is 209,613.\footnote{We collect versions of articles in other language editions as needed for analysis. Decisions about which data and categories to focus on were made in consultation with researchers at the Wikimedia Foundation.}

We briefly summarize our methods for inferring race and gender and provide additional details in \Aref{sec:app_wikidata}.
Identity traits are fluid, difficult to operationalize, and depend on social context \cite{Bucholtz2005,hanna2020towards}.
Our goal is to identify \textit{observed} gender and race as perceived by Wikipedia editors who assigned article metadata or readers who may view them, as opposed to assuming ground-truth values \cite{roth2016multiple}. Thus, we derive race and gender directly from Wikipedia articles and associated metadata. We treat identity as fixed at the time of data collection and caution that identity traits inferred in this work may not be correct in other time-periods or contexts.

We primarily infer gender from Wikidata---a crowd-sourced  database corresponding to Wikipedia pages \cite{vrandevcic2014wikidata}, focusing on 5 genders common in Wikidata. We use cis.~men as the comparison group and separately identify matches from this group for each other gender.


As race is a social construct with no global definition,\footnote{\url{https://unstats.un.org/unsd/demographic/sconcerns/popchar/popcharmethods.htm}}
and given that our starting point for data collection is English Wikipedia, we focus on biographies of American people and use commonly selected race/ethnicity categories from the U.S census: Hispanic/Latinx, African American, and Asian.
We use these categories because we expect the predominately U.S.-based English Wikipedia editors to be familiar with them, but we do not suggest that they are naturally occurring or meaningful in other contexts. To identify pages for each target group, we primarily use category information, identifying articles for each racial group as ones containing categories with the terms ``American'' and [“Asian”, “African”, or “Hispanic/Latinx”] or names of specific countries. We acknowledge that our use of the term ``race'' is an oversimplification of the distinctions between articles in our corpus. Nevertheless, we believe it reflects the perceptions that Wikipedia users may hold.

Finally, a natural choice for comparisons would be articles about white/Caucasian Americans. However, we encountered the obstacle of ``markedness'': while articles about racial minorities are often explicitly marked as such, whiteness is assumed so generic that it is rarely explicitly marked \cite{brekhus1998sociology,trechter2001introduction}. We see this in our data: Barack Obama's article has the category ``African American United States senators'', whereas George W. Bush's article has the category ``Governors of Texas'', not ``White Governors of Texas''. Thus, we consider markedness itself a social indicator and use ``unmarked'' articles as candidate comparisons: all pages that contain an ``American'' category, but do not contain a category or Wikidata entry property indicative of non-white race.\footnote{(including Middle Eastern, Native American, and Pacific Islander) We also exclude football/basketball players, as these articles were very often unmarked, as were articles about jazz musicians.
We suggest further investigation for future work.} In manually reviewing the comparison corpus, based on outside Wikipedia sources and pictures, we estimated that it consists of ~90\% Caucasian/white people.

\section{Evaluation Results}
\label{sec:results}

\Fref{fig:random_eval} reports evaluation results for 100 \textit{article-sampling} and \textit{category-sampling} simulations.
In addition to the described matching algorithms, we show the results of randomly sampling a comparison group. All evaluation metrics measure differences between the target and comparison groups: lower values indicate a better match. 
Throughout all evaluations, except where explicitly noted, we do not exclude weak matches in order to retain comparable target sets. Exclusion can result in different target articles being discarded under different matching approaches. 

All methods perform better than random in reducing covariate imbalance, and \textsc{Pivot-Slope TF-IDF} best reduces the percentage of highly imbalanced categories (\%$\text{SMD}{>} 0.1$). In the category-sampling simulations (bottom), which better simulate having a target group with a particular trait in common, all methods also perform better than random in the text-based metrics (PLO and KL), and \textsc{Pivot-Slope TF-IDF} performs best overall. In article-sampling simulations (top), random provides a strong baseline. This is unsurprising, as randomly chosen groups of 1000 articles are unlikely to differ greatly. Nevertheless, \textsc{Pivot-Slope TF-IDF} outperforms random on covariate balancing and the text-based metrics.

\begin{figure}
    \centering
    \includegraphics[width=0.9\columnwidth]{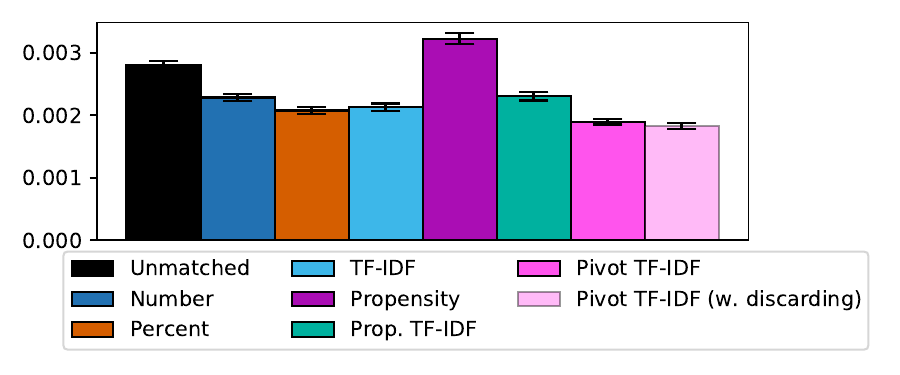}
    \Description[Bar graphs of SMD for African American people's articles]{Bar graphs of SMD evaluation metric for eight methods used to create matches for articles about African American people}
    \caption{SMD between articles about African American people and matched comparisons, averaged across categories with 99\% confidence intervals. Lower scores indicate a better match. ``(w. discarding)'' indicates SMD after weak matches are discarded.}
    \label{fig:african_american_eval}
\end{figure}

\begin{figure*}
    \centering
    \includegraphics[width=0.99 \textwidth]{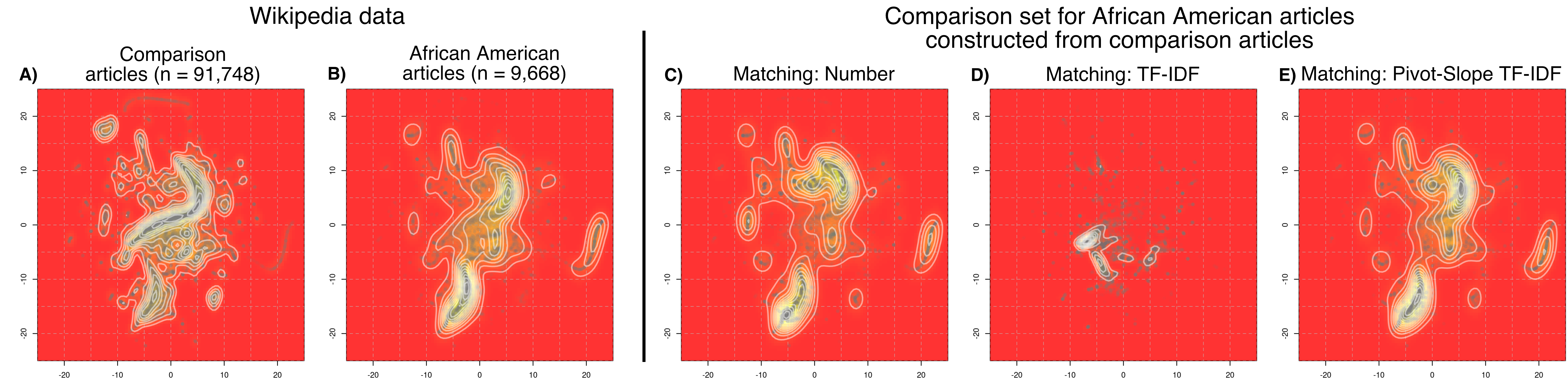}
    \Description[UMAP for African American people's articles]{UMAP visualizations showing that categories of matched articles have a more similar visualized shape to categories associated with African American articles than ones of unmatched articles}
    \caption{UMAP visualizations of TF-IDF weighted categories, where
    each point represents an article, and the red-to-yellow coloring depicts the low-to-high density of articles emphasized by white contours. (A) and (B) shows category distributions in African American and (unmatched) comparison articles respectively. (C) - (E) show category distributions in comparison sets generated by different matching methods. \textsc{Percent} and propensity methods (not shown for brevity) perform strictly worse.
    \textsc{Pivot-Slope TF-IDF} (D) results in the comparison set with the most similar category distribution as the target set (B).  }
    \label{fig:umap_african_american}
\end{figure*}

As expected, \textsc{Number} exhibits bias towards articles with more categories, while \textsc{Percent} and \textsc{TF-IDF} exhibit bias towards articles with fewer categories, resulting in worse performance than random over the the number of categories (\# Cat.) metric (\Fref{fig:random_eval} reports absolute values). These differences are also reflected in text length, as articles with more categories also tend to be longer. In category-sampling, pivot-slope normalization corrects for this length bias and outperforms random. In article-sampling, while \textsc{Pivot-Slope TF-IDF} does outperform other metrics, the random sampling exhibits the least category number and text length bias. However, as mentioned, random is a strong baseline in this setting.

In \Fref{fig:african_american_eval} we present an \textit{attribute-specific} evaluation: SMD averaged across categories between articles about African American people and comparisons, under different matching methods (evaluations for other target groups in \Aref{sec:app_eval} show similar patterns). As in \Fref{fig:random_eval}, we generally do not exclude weak matches in order to have directly comparable target sets, though we do report SMD after this exclusion for \textsc{Pivot-Slope TF-IDF} in order to accurately reflect SMD in our final analysis data. We further note that if we discard weak matches for all methods, \textsc{Pivot-Slope TF-IDF} results in the least amount of discarded data.
With the exception of \textsc{Propensity}, all methods improve covariate balance as compared to not using matching, and \textsc{Pivot-Slope TF-IDF} performs best.

To supplement the quantitative results, we additionally provide qualitative results via UMAP \citep{mcinnes2018umap}.
Given a matrix $X \in \mathcal{R}^{n * k}$ with $n$ rows (i.e., articles) and $k$ columns (i.e., categories), UMAP nonlinearly maps each row into a two-dimensional space 
that preserves nearest-neighbor geometry.
UMAP is often preferred over other nonlinear dimension reduction methods for its effectiveness in visualizing the data and assessing clustering structure, e.g., in text-analyses \citep{ochigame2021search,bolukbasi2021interpretability} and genomics \citep{becht2019dimensionality}.
We set $X$ to be the matrix of TF-IDF weighted categories for all methods to obtain a global set of $n$ coordinates (one per article). Details are provided in \Aref{sec:app_umap}.

\Fref{fig:umap_african_american} shows UMAP visualizations for African Americans. Without matching, the set of all comparison articles (A) has a distinctly different distribution of associated categories than articles about African American people (B), which motivates our work. Of the matching methods,
\textsc{Pivot-Slope TF-IDF} (E) produces a comparison set with the most similar distribution of associated categories as those from articles of African American people. While \textsc{Number} (C) does also produce a similar category distribution, \Fref{fig:random_eval} demonstrates that this method favors articles with more categories. \textsc{TF-IDF} (D) performs particularly poorly.  As this method overly-favors articles with too few categories, comparison articles with few categories appear in a disproportionate number of matches. 


\section{Analysis}
\label{sec:analysis}

Finally, we use \textsc{Pivot-slope TF-IDF} matching to facilitate an examination of possible biases and content gaps along gender and race dimensions. We compute several metrics over English articles drawn from prior work \cite{wagner2015s} including article length, language availability, edit count, article age, log-odds scores of words in the articles \cite{monroe2008fightin}, and percent of article devoted to common sections. Additionally, for the top 10 most edited languages (English, French, Arabic, Russian, Japanese, Italian, Spanish, German, Portuguese, and Chinese), for all target-comparison pairs where both members of the pair are available in the language, we compare article lengths and normalized section lengths.
While there are numerous metrics that could be examined over these data sets, we choose metrics that are likely to reveal content biases found in prior work. For example, language availability and article length reveal how Wikipedia might provide higher volume of information for some groups of people than others. Differences in section structure are reflective of article quality \citep{Piccardi2018} and likely to reflect previously observed content biases, e.g., if articles about women discuss personal relationships more frequently than ones for men, we would expect them to have longer ``Personal Life'' sections \cite{graells2015first,wagner2016women,adams2019counts}. Article age and edit counts can provide some insight into whether observed differences are likely reflective of editing habits or other factors.

In general, we compute if differences in statistics between target and comparison groups are significant using paired t-tests for continuous values (e.g. average article length) and McNemar’s test for dichotomous values (e.g. if article is available in German or not). We use Benjaminin Hochberg multiple hypothesis correction for metrics with many hypotheses (e.g. if target or comparison articles are more likely to be available in each of 50 languages). Due to space limitations, we discuss a subset of results and provide a webpage with complete metrics.\footnote{
\url{https://anjalief.github.io/wikipedia_bias_viz/}
}

\textbf{Matching Reduces Data Confounds}
First, we revisit our motivation by considering how results differ without matching. Table \ref{tab:gender_odds} presents the words most associated with biography pages about cis. men and women calculated using log-odds \cite{monroe2008fightin}. As shown in previous work, without matching, words highly associated with men include many sports terms, which suggests that directly comparing these biographies could capture athlete/non-athlete differences rather than man/woman differences. After matching, these sports terms are replaced by overtly-gendered terms like ``himself'' and ``wife'', showing that matching helps isolate gender as the primary variable of difference between the two corpora. Beyond the top log-odds scores, sports terms do occur, but they tend to be more specific and represented on both sides, for example ``WTA'' is women-associated and ``NBA'' is men-associated.


\begin{table}
    \centering
    \begin{tabular}{ll}
    Unmatched (M) & he/He, his/His, \textbf{season}, him, \textbf{League}, \textbf{club} \\
    Unmatched (W) & her/Her, she/She, women, actress, husband \\
    \hline
    Matched (M) & he/He, his/His, him, himself, wife, Men \\
    Matched (W) & her/Her, she/She, women, husband, female \\
    \end{tabular}
    \caption{Log-odds scores between cis. men and women pages ordered most-to-least polar from left to right. Matching reduces sports terms (bold) in favor of overtly gendered terms.}
    \label{tab:gender_odds}
\end{table}


\Tref{tab:race_lengths_unmatched} shows the results of comparing article lengths between racial subgroups and the entire candidate comparison group, rather comparing with the subset of matched articles. Articles for all racial subgroups are significantly longer than comparison articles.

However, after matching (\Tref{tab:general_statistics}), we do not find significant differences between matched comparison articles and articles about African American and Hispanic/Latinx American people.
Instead, we find that articles about Asian American people are typically shorter than comparison articles.

\begin{table}
    \centering
    \begin{tabular}{cccc}
    \hline
    & Target & Comparison  \\
    \hline
    \hline
    African American & 902.0 & 711.4  \\
    Asian American & 737.5 & 711.4 \\
    Hispanic/Latinx American  & 972.5 & 711.4 \\
    \hline
    \end{tabular}
    \caption{Average article lengths without matching. All target sets appear significantly ($\text{p}{<}0.05$) longer than comparisons.}
    \label{tab:race_lengths_unmatched}
\end{table}

\begin{table*}
    \centering
    \begin{tabular}{lccccccccc}
    & {\#Pairs analyzed} & \multicolumn{2}{c}{Article Lengths} & \multicolumn{2}{c}{Edit History} &  \multicolumn{2}{c}{Article Age} & \multicolumn{2}{c}{\# of Languages} \\ 
    \cmidrule(lr){3-4} \cmidrule(lr){5-6} \cmidrule(lr){7-8} \cmidrule(lr){9-10}
    & & Target & Comparison & Target & Comparison  & Target & Comparison & Target & Comparison \\
    \hline
    \hline
    African Amer. & 8,404 & 942.9 & 959.2 & 243.4 & 245.8  & \textbf{128.5} & 136.2 & \textbf{6.2} & 6.8 \\
    Asian Amer. & 3,473 & \textbf{792.3} & 854.1 & 193.2 & 198.5 & \textbf{123.2} & 130.3 & \textbf{6.0} & 7.1 \\
    Hisp./Latinx Amer. & 3,813 & 1017.2 & 1026.8  & 293.4 & 277.8 & \textbf{130.0} & 137.4 & 7.5	& 7.6 \\
    \hline
    Non-Binary & 127 & 1086.5 & 914.9 & 374.0 & 189.1 & \textbf{95.0} & 119.7 & 7.8 & \textbf{5.9} \\
    Cis. women & 64,828 & \textbf{668.9} & 792.4 & \textbf{126.1} & 147.2 & \textbf{110.6} & 128.7 & \textbf{5.4} & 6.1 \\
    Trans. women & 134 & 1115.3 & 837.1 & 270.5 & \textbf{151.6} & \textbf{119.6} & 135.3 & 8.3 & \textbf{5.52}\\
    Trans. men & 53 & 652.7 & 870.9 & 118.2 & 172.0 & \textbf{97.0} & 125.7 & 3.9 & 5.8 \\
    \hline
    \end{tabular}
    \caption{Averaged statistics for articles in each target group and matched comparisons, where matching is conducted with Pivot-Slope TF-IDF. For statistically significant differences between target/comparison ($\text{p} {<} 0.05$) the smaller value is in bold.}
    \label{tab:general_statistics}
\end{table*}

\textbf{Metrics Reveal Content Imbalances}
We present several high-level statistics in \Tref{tab:general_statistics}, which can identify possible content gaps and sets of articles that may benefit from additional editing.
Articles about Asian Americans and cis.~women tend to be shorter than comparison articles. Shorter articles can indicate that articles were written less carefully, e.g., editors may have spent less time uncovering details to include in the article,
though paradoxically, as editors often spend time cutting `bloated' articles, shorter articles could also be reflective of increased editor attention.
They can also indicate that information is less carefully presented---in examining matched pairs, we often found that information in Asian American peoples' articles was displayed tables, whereas information in comparison articles was presented in descriptive paragraphs.

The center columns in \Tref{tab:general_statistics} provide additional context by comparing the average edit count and age (in months; at the time that edit data was collected) for each article.\footnote{Edit data was collected in Sept. 2020, 6 months after the original data; a few articles which had been deleted or had URL changes between collection times are not included.} Notably, all articles about gender and racial minorities were written more recently than matched comparisons, which could reflect growing awareness of biases on Wikipedia and corrective efforts.\footnote{Example: \url{https://meta.wikimedia.org/wiki/Gender_gap}


} Furthermore, articles about cis.~women do have fewer edits than comparisons, while articles about Asian Americans do not have significantly fewer edits (even though both were written more recently than comparisons).


While prior work has suggested that articles about cis.~women tend to be longer and more prominent than articles about cis.~men, our work has the opposite finding \cite{graells2015first,reagle2011gender,wagner2015s,young2016s}. There are several differences in our work: use of matching, discarding incomplete ``stub'' articles, focus on ``Living People'', and consideration of gender as non-binary. Also, Wikipedia is constantly changing, and prior work identifying missing pages of women on Wikipedia has caused editors to create them \cite{reagle2011gender}. However, our work suggests that a disparity between the quality articles about cis.~men and women still exists. 
Given the differences in edit counts, greater editor attention to articles about cis.~women could reduce this gap.
In contrast, more investigation of disparity in articles about Asian American people is needed, since there is not a significant difference in the number of edits per article. Edit counts offer an overly simple view of the editing process, and a more in-depth analysis,
could offer more insights. Nevertheless, our results support work contradicting the ``model minority myth'' \cite{chou2015myth} in suggesting that Asian Americans are not exempt from prejudice and racial disparities.

\textbf{Non-English Statistics Reveal ``Local Heros''}
We next consider other language editions. From the rightmost columns in \Tref{tab:general_statistics}, articles about African Americans, Asian Americans, and cis.~women are typically available in fewer languages than comparisons. In contrast, articles about non-binary people and transgender women are available in significantly more languages (discussed below). Language availability can indicate underrepresentation---a user searching in non-English  Wikipedia editions is less likely to find biography pages of African Americans than other Americans.

When we examine what percentage of target vs. comparison articles are available in each language, we find that disparities occur broadly across many languages. Articles about African Americans are significantly more likely to have versions in Haitian, Yoruba, Swahili, Punjabi, and Ido, and less likely in 47 other languages. Articles about Asian Americans are significantly more likely to have versions in Hindi, Punjabi, Chinese, Tagalog, Tamil, and Thai and less likely in 42 other languages. Similarly, articles about cis.~women are more available in 18 languages and less available in 63 languages. For reference, articles about Latinx/Hispanic people, where we do not see a significant difference in overall language availability (\Tref{tab:general_statistics}), are significantly more likely to have versions in Spanish and Haitian and less likely in 8 other languages.
These results support prior work on ``local heros'' showing that a person's biography is more likely to be available in languages common to the person's nationality \cite{eom2015interactions,callahan2011cultural,hecht2010tower}. Our results show that pattern holds for a person's ethnicity and background, beyond current nationality.
These results also show that reducing these observed language availability gaps
requires substantial effort, as it requires adding articles in a broad variety of languages, not just a few.

We can examine additional information gaps by considering how the lengths of the same articles differ between languages.
While article lengths can differ because of a number of factors,
length differences in one language that do not exist in another language indicate a content gap that can be addressed through additional editing---we know that the disparity does not occur because of factors independent of Wikipedia because it does not exist in the other language.
We note two observed disparities. For 240 articles about African American people, both the article and its match are available in Chinese. The articles about African Americans are significantly shorter than their matches in Chinese (target length: 25.7, comparison length: 36.0, p-val: 0.044), but not in English (t: 2,725.3, c: 2,500.4, p: 0.22). Similarly, for the 912 matched pairs available in Spanish, the articles about African Americans are significantly shorter in Spanish (t: 659.8, c: 823.9, p: 0.003), but not English (t: 1,639.2, c: 1,544.1, p: 0.18).

\textbf{Statistics Align with Social Theories}
\label{sec:glass_ceiling}
While our goal is not a thorough investigation of social theories, we briefly discuss``the glass ceiling effect'' and ``othering'', to demonstrate how our methodology can support this line of research, and how existing theory can provide insight into our data.

\citet{wagner2016women} show that women on Wikipedia tend to be more notable than men and suggest a subtle glass ceiling: there is a higher bar for women to have a biography article than for men. We can find evidence of glass ceiling effects by comparing article availability and length across languages. Specifically, articles about African American people are significantly less available in German (t: 29.87\% c: 31.75\%	p-val: 0.0041). However, for the subset of 1,399 pairs, for which both the matched target and comparison articles are available in German, the English target articles are significantly longer than comparisons (t: 1,406.43, c: 1,291.38 p-val: 0.02). While numerous factors can account for why an article may not be available in a particular language, one possible reason for the difference in German is that articles about African American people only exist for particularly noteworthy people with long articles detailing many accomplishments, whereas a broader variety of articles about other Americans are available.

Additionally, our results show that the biography pages of transgender women and non-binary people tend to be longer  and available in more languages than matched comparison articles (\Tref{tab:general_statistics}; length differences are not statistically significant, likely due to small data size).
%
While this could indicate a glass ceiling effect; another factor contributing to article length is that in these pages, the focal person's gender identity is often highlighted.
Both non-binary people and transgender women tend to have a higher percentage of their article devoted to the ``Personal Life'' section (non-binary: 8.54\%, comparison: 2.70\%	p-val: 1.97E-05), (transgender women: 7.32\%, comparison: 1.92\%, p-val: 4.78E-05).
In examining these pages, personal life sections often focus on gender identity.
The implication that gender identity is a noteworthy trait for just these groups is possibly indicative of ``othering'', where individuals are distinguished or labeled as not fitting a societal norm, which often occurs in the context of gender or sexual orientation \cite{Mountz2009,nadal2012interpersonal}. We 
leave more in-depth exploration for future work.



\begin{table}
    \centering
    \begin{tabular}{lcccc}
    \hline
    & \#Pairs & Length & Lang.   \\
    \hline
    \hline
    vs. unmarked Amer. women & 2,930 & -41.97  & \textbf{1.29} \\
    vs. African Amer. men & 1,921 & -31.39  &  0.18\\
    vs. unmarked Amer. men & 2,132 & -54.01 & \textbf{-0.48} \\
    \end{tabular}
    \caption{Differences in average article lengths and language availability (comparison - target), for African American cis. women vs. comparisons. Significant results are bolded.}
    \label{tab:intersectional_stats}
    \vspace{-5mm}
\end{table}

\textbf{Intersectionality Uncovers Differing Trends}
Literature on \textit{intersectionality} has shown that discrimination cannot be properly studied along a single axis. For example, focus on race tends to highlight gender privileged people (e.g. Black men), while focus on gender tends to highlight race privileged people (e.g. white women), which further marginalizes people who face discrimination along multiple axes (e.g. Black women) \cite{crenshaw1989demarginalizing,schlesinger2017intersectional,rankin2019straighten}.
Although our work only focuses on two dimensions, which is not sufficient for representing identity, we can expand on the prior single-dimensional analyses by considering intersected dimensions. We focus on African American cis. women for alignment with prior work \cite{crenshaw1989demarginalizing} and use 3 comparison groups: unmarked American cis. women, African American cis. men, and unmarked American cis. men (Table \ref{tab:intersectional_stats}).
Notably, articles about African American cis. women are available in significantly fewer languages than unmarked American cis. women, even though these articles are often longer (though not significantly).
In contrast, articles about African American cis.~women are available in more languages than articles about unmarked American cis.~men.
These results suggest that African American women tend to have more global notoriety then comparable American men, possibly a ``glass ceiling'' or a ``othering'' effect. However, African American  women do not receive as much global recognition on Wikipedia as comparable American women.
This result supports the theory that focusing on gender without considering other demographics can mask marginalization and fails to consider that some women face more discrimination than others.

\textbf{Conclusion} 
We present a method 
that can be used to construct controlled corpora for documents with sparse metadata. As demonstrated in our  analysis, this methodology can help identify systemic differences between sets of articles, facilitate analysis of specific social theories, and provide guidance for reducing bias in corpora.

\begin{acks}
We thank Martin Gerlach, Lucille Njoo, Xinru Yan, Michael Miller Yoder, and Leila Zia for their helpful feedback.
A.F. would like to acknowledge support from a Google PhD Fellowship. This material is based upon work supported by the National Science Foundation under Grant No.~IIS2040926 and the DARPA CMO under Contract No.~HR001120C0124. 
\end{acks}
\bibliographystyle{ACM-Reference-Format}
\bibliography{references.bib}

\newpage
\clearpage
\pagebreak

\appendix

\section{Experimental setup}
\label{sec:app_experimental_setup}

As described in \Sref{sec:method}, in constructing \textsc{pivot-slope TF-IDF} vectors, we set the pivot to 9.3 (avg. number of categories). We constructed two development sets for tuning the slope:  1000 randomly sampled pages and 500 random pages from each of 10 randomly sampled categories. We set the slope to the value between 0 and 1 (0.1 increments) that minimized the metrics described in \Sref{sec:experiments} (slope=0.3) over the tuning set. We excluded the biography pages used for tuning during evaluation. 
For all methods, we perform matching with replacement, and we limit the number of times an article can appear in $\mathcal{C}$ (e.g., the number of replacements) to 10; we observe no notable differences in results when replacement is unlimited.

In computing the Polar Log-odds (PLO) evaluation metric, we take the absolute value of log-odds scores for all target and comparison words, and compute the mean and standard deviation for the 200 most polar words. 

In computing the KL Divergence evaluation metric, we train an LDA model with 100 topics across all articles in the corpus \cite{blei2003latent}. After matching, we average the topic vector for each article in the comparison and the target group, using $\frac{1}{1000}$ additive smoothing to avoid 0 probabilities, and then normalize these vectors into valid probability distributions. We then compute the KL-divergence between the target and comparison topic vectors.

\section{Details and data statistics for gender and race identification}
\label{sec:app_wikidata}

\Tref{tab:data_sizes} reports data sizes. We use the following procedures to infer gender associated with Wikipedia articles:

\textbf{Transgender men/women} Articles whose Wikidata entry contains Q\_TRANSGENDER\_MALE/\_FEMALE.

\textbf{Non-binary, gender-fluid, or gender-queer; termed ``non-binary''} Articles whose Wikidata entry contains Q\_NONBINARY or Q\_GENDER\_FLUID.
Also, pages with a category containing "-binary" or "Genderqueer", as we found that some pages with non-binary gender indicators had binary gender properties in Wikidata.\footnote{e.g.~\url{https://en.wikipedia.org/wiki/Laganja\_Estranja}}

\textbf{cis. women/men} Articles whose Wikidata entries contain the property Q\_FEMALE/Q\_MALE. We also assigned 38,955 articles for which we could not identify Wikidata entries nor any indication of non-binary gender as cis.~man or cis. woman based on the most common pronouns used in the page. For pages that did have Wikidata entries, pronoun-inferred gender aligned with their Wikidata gender in 98.0\% of cases.\footnote{There can be errors in Wikidata \cite{Heindorf2019}}  We exclude cis.~men with categories containing the keyword ``LGBT'' when matching transgender and non-binary people to them.

As racial definitions depend on social context, we focus on racial categories commonly used in the U.S.\footnote{e.g. United Nation statistics define ethnics/national groups as ``dependent upon individual national circumstances'' and European Union directives reject theories that attempt to define separate human races \url{https://eur-lex.europa.eu/LexUriServ/LexUriServ.do?uri=CELEX:32000L0043:en:HTML}
\url{https://unstats.un.org/unsd/demographic/sconcerns/popchar/popcharmethods.htm}}
To infer race associated with Wikipedia articles: we identify articles for each racial group as ones containing categories with the terms ``American'' and [“Asian”, “African”, or “Latinx/Hispanic”].\footnote{
While the census considers Hispanic/Latinx an ethnicity, surveys suggest that 2/3 of Hispanic people consider it part of their racial background 
\url{https://www.pewresearch.org/fact-tank/2015/06/15/is-being-hispanic-a-matter-of-race-ethnicity-or-both/}
}
We also include categories containing ``American'' and the name of a country in Asia, Africa, or Latin American (including the Caribbean),\footnote{We identify country lists from \url{worldometers.info}} and we filter categories through further heuristics, e.g.~discarding ones containing ``expatriates'' and ``American descent''.
Our final category set includes ones like ``American academics of Mexican descent'', and excludes ``American expatriates in Hong Kong''.
 In general, we do not force racial categories to be exclusive, e.g.~410 pages are considered both African American and Hispanic/Latinx (we do enforce that people in any target corpora cannot be in the comparison corpora). 

\begin{table}
    \centering
    \begin{tabular}{lcc}
    \hline
    Group & Pre-match & Final \\
    \hline
    \hline
    African American & 9,668 & 8,404 \\
    Asian American & 4,728 & 3,473 \\
    Hispanic/Latinx American & 4,483 & 3,813 \\
    Unmarked American (comparison) & 93,486 & - \\
    \hline
    Non-Binary & 200 & 127 \\
    Cisgender women & 108,915 & 64,828 \\
    Transgender women & 261 & 134 \\
    Transgender men & 85 & 53 \\
    Cisgender men (comparison) & 331,484 & - \\
    \hline
    \end{tabular}
    \caption{Data set sizes for analysis corpora. ``Final'' column indicates the target/comparison sizes after corpora are matched using \textsc{Pivot-Slope TD-IDF} and matches with $<2$ categories in common are discarded.}
    \label{tab:data_sizes}
\end{table}

We validate our category-based approach by comparing the set of articles that we identify as describing African American people with the Wikidata ``ethnic group'' property. We cannot use this property for all target groups, as it was only populated for 3.4\% of articles in our data set and is largely unused for relevant values other than ``African American.''
Of the 9,668 people that our category method identifies as African American, we were able to match 5,776 of them to ethnicity information in Wikidata. Of these 5,776 pages, our method exhibited  98.5\% precision and 69.0\% recall. Here, precision is more important than recall, as low recall implies we are analyzing a smaller data set than we may have otherwise, while low precision implies we are analyzing the wrong data.

\begin{figure*}
     \begin{subfigure}[b]{0.33\textwidth}
         \centering
        \includegraphics[width=\textwidth]{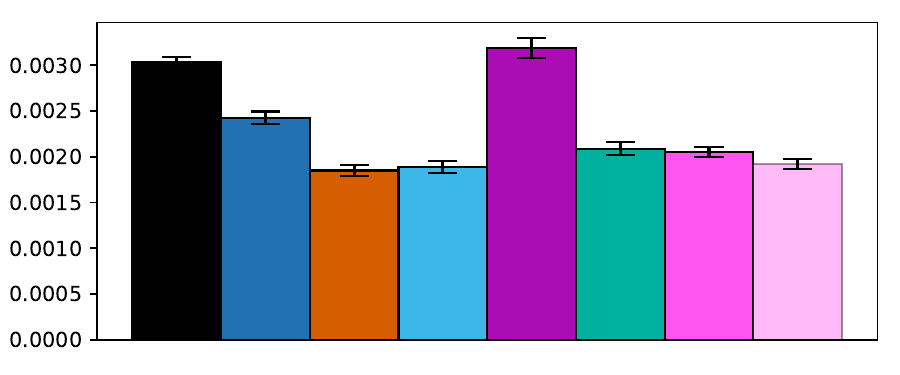}
         \caption{Asian American People}
     \end{subfigure}
     \begin{subfigure}[b]{0.33\textwidth}
         \centering
        \includegraphics[width=\textwidth]{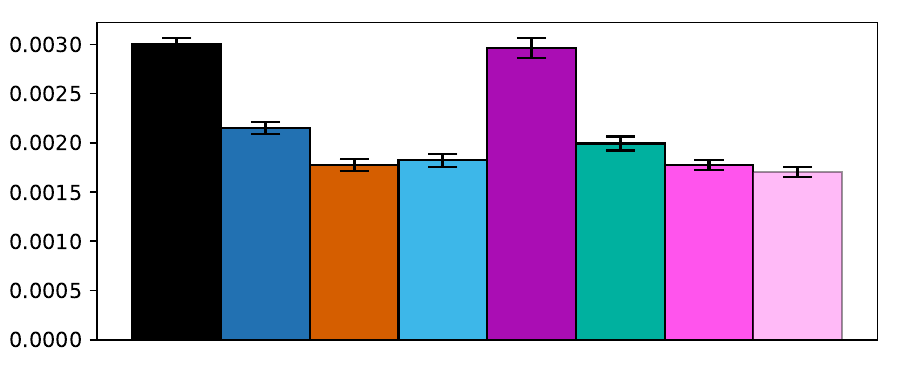}
         \caption{Hispanic/Latinx American people}
     \end{subfigure}
     \begin{subfigure}[b]{0.33\textwidth}
         \centering
        \includegraphics[width=\textwidth]{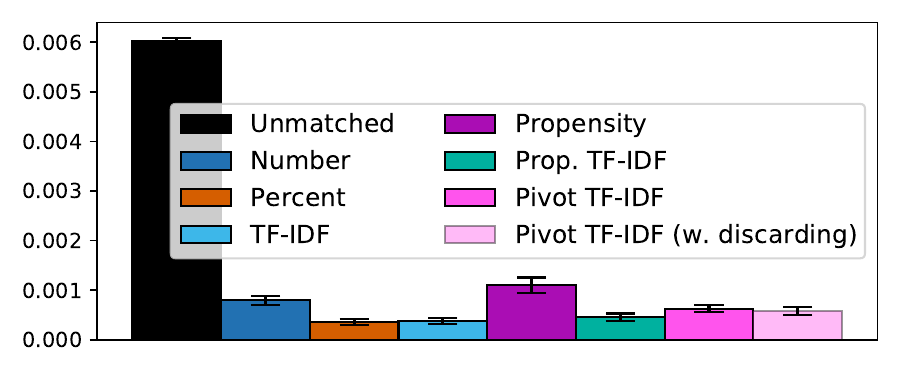}
         \caption{Transgender men}
     \end{subfigure}

     \begin{subfigure}[b]{0.33\textwidth}
         \centering
        \includegraphics[width=\textwidth]{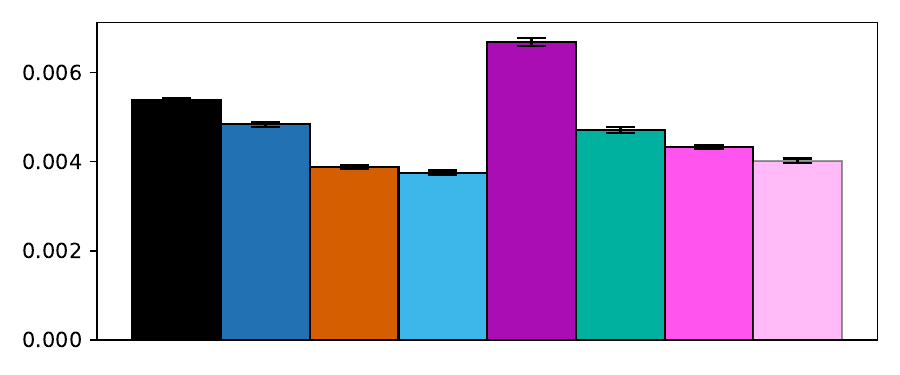}
         \caption{Cisgender women}
     \end{subfigure}
     \begin{subfigure}[b]{0.33\textwidth}
         \centering
        \includegraphics[width=\textwidth]{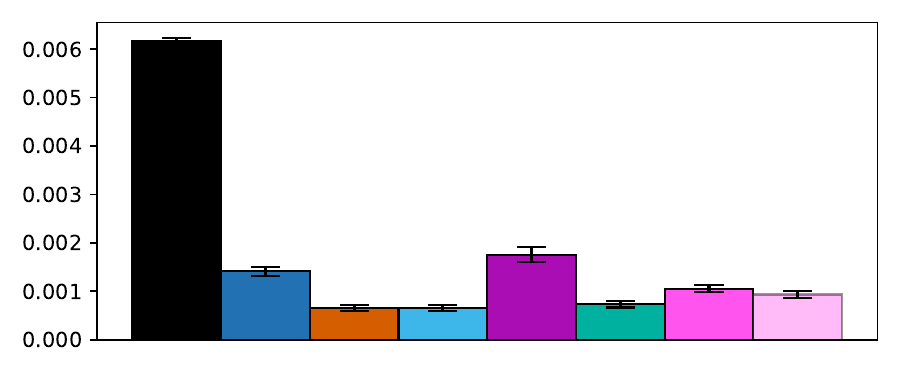}
         \caption{Transgender women}
     \end{subfigure}
     \begin{subfigure}[b]{0.33\textwidth}
         \centering
        \includegraphics[width=\textwidth]{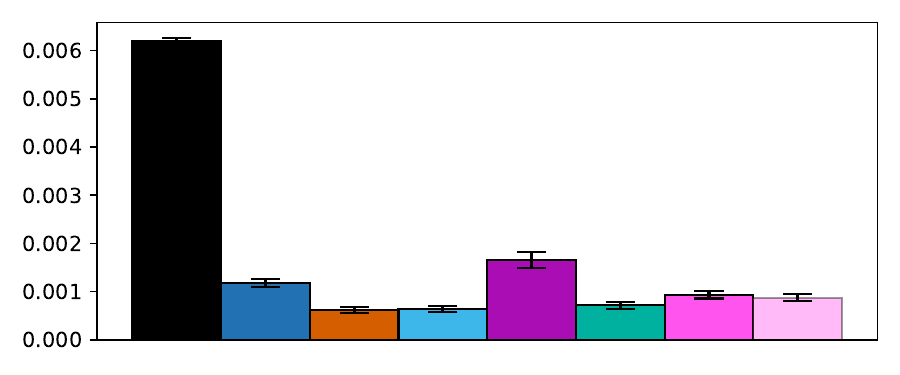}
         \caption{Non-binary people}
     \end{subfigure}

     \begin{subfigure}[b]{0.33\textwidth}
         \centering
        \includegraphics[width=\textwidth]{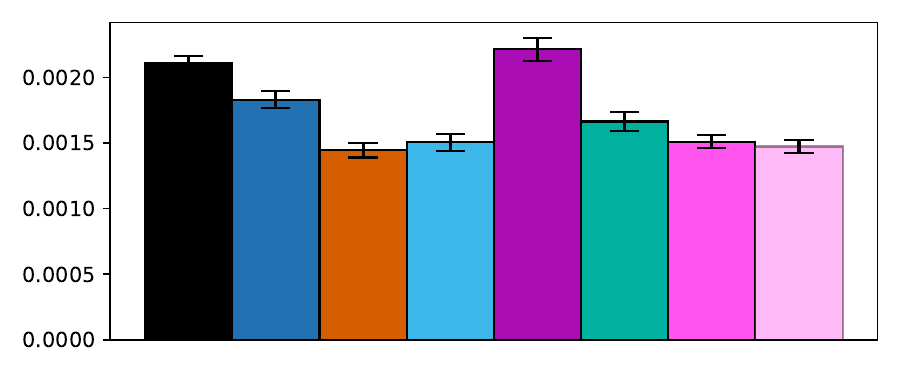}
         \caption{African American cis. women compared with unmarked cis. women}
     \end{subfigure}
     \begin{subfigure}[b]{0.33\textwidth}
         \centering
        \includegraphics[width=\textwidth]{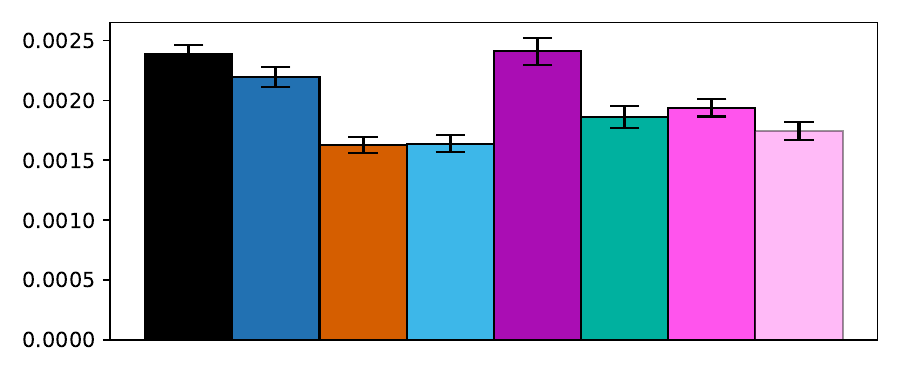}
         \caption{African American cis. women compared with African American cis. men}
     \end{subfigure}
     \begin{subfigure}[b]{0.33\textwidth}
         \centering
        \includegraphics[width=\textwidth]{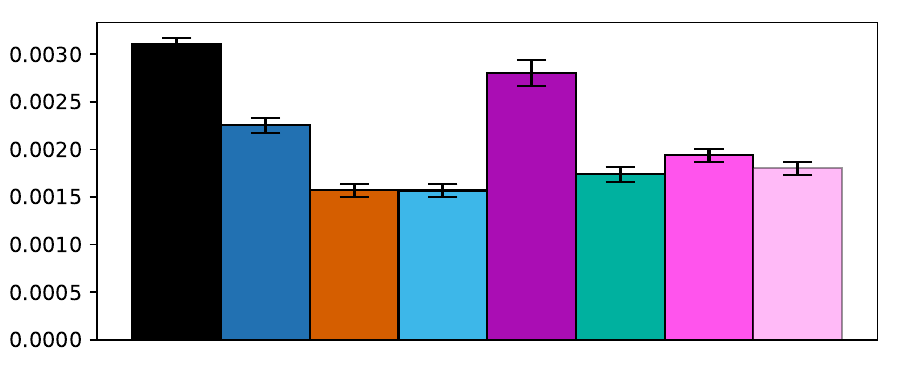}
         \caption{African American cis. women compared with unmarked cis. men}
     \end{subfigure}
    \Description[Bar graphs of all SMD]{9 bar graphs showing SMD evaluation metric for 9 target groups, each which with 8 bars representing different matching method}
    \caption{Standardized mean differences averaged across categories for matched articles for all additional target groups. The legend in (c) applies to all figures.}
    \label{fig:app_match_eval}
\end{figure*}

\begin{figure*}
    \centering
    \includegraphics[width=0.99 \textwidth]{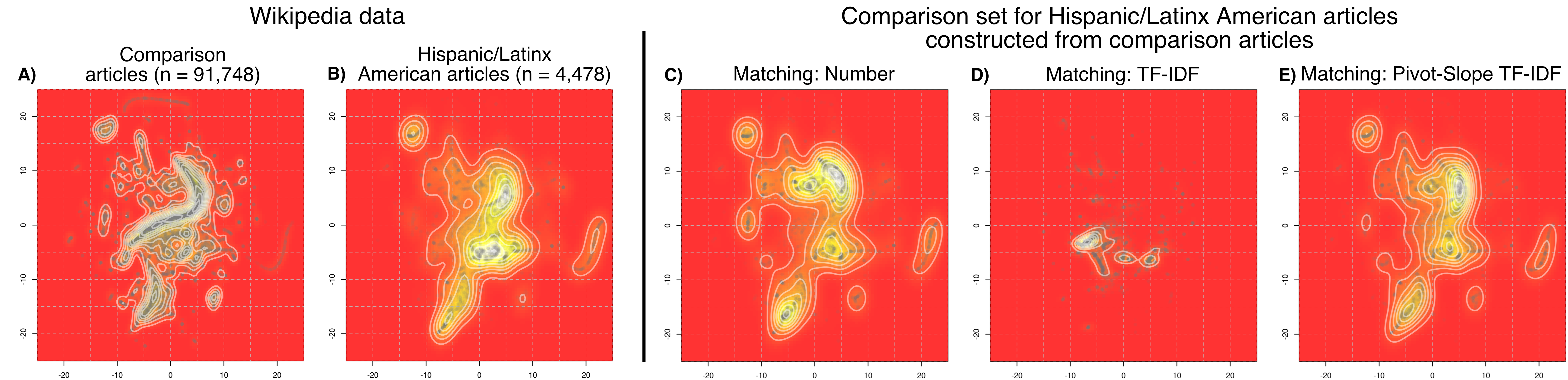}
    \caption{UMAP visualizations of TF-IDF weighted categories associated with articles about Hispanic/Latinx Americans and comparison people, analogous to \Fref{fig:umap_african_american}. Note that the distribution of categories
    infered from (B) are different from those corresponding to African Americans in
    \Fref{fig:umap_african_american}(B).}
    \label{fig:umap_latinx_american}
\end{figure*}

\section{UMAP details and justification}
\label{sec:app_umap}
The UMAPs were run on the $X \in \mathcal{R}^{n * k}$ using the TF-IDF weighted categories via 
\texttt{uwot::umap} function in \texttt{R},
where it is applied on the leading-50 principal scores, and using Euclidean distance to measure distance. Additionally, we use \texttt{min\_dist=0.1}, \texttt{spread=2}, and \texttt{n\_neighbors=100}. 
We apply UMAP on the TF-IDF weighted categories over the one-hot encoded matrix (i.e., the binary matrix annotating which categories are associated with with article) since numerous 
work have shown that the nearest-neighbors of the matrix of TF-IDF weighted attributes are more meaningful
than those in the one-hot encoded matrix (see \cite{qaiser2018text} and \cite{trstenjak2014knn} for example).

\section{Additional Evaluation Results}
\label{sec:app_eval}

\Fref{fig:app_match_eval} shows SMD values for all target groups analyzed in this work akin to \Fref{fig:african_american_eval}. In general, matching improves SMD as compared to not matching. \textsc{Pivot-Slope TF-IDF} generally performs the best. In some cases, \textsc{TF-IDF} and \textsc{Percent} do reduce SMD more than \textsc{Pivot-Slope TF-IDF}. However, in \Sref{sec:results} we discuss the limitations of these methods, that they heavily favor articles with fewer categories.
Figure \ref{fig:umap_latinx_american} shows the UMAP for the articles corresponding to Hispanic/Latinx Americans and the comparison sets formed by different matching methods, analogous to \Fref{fig:umap_african_american}. Similar to our discussion in \Sref{sec:results}, the \textsc{Pivot-Slope TF-IDF} matches output a distribution of categories most-similar to the target set.

\end{document}